\documentclass{article}
\usepackage{spconf,amsmath,graphicx}
\usepackage{fancyhdr}
 \usepackage{epsfig,amssymb,amsmath,multirow,boldline,graphicx,makecell,hyperref,subcaption,textcomp}
\usepackage{kotex}
\usepackage{float}
\usepackage{multirow}
\usepackage{xcolor}

\title{Noise-Tolerant Audio-Visual Online Person Verification\\ Using an Attention-based Neural Network Fusion\\
}

\name{Suwon Shon, Tae-Hyun Oh, James Glass}

\address{MIT Computer Science and Artificial Intelligence Laboratory, Cambridge, MA, USA\\
{\small \tt \{swshon,glass\}@mit.edu, taehyun@csail.mit.edu} }
\begin{document}
\ninept
\maketitle
\begin{abstract}
In this paper, we present a multi-modal online person verification system using both speech and visual signals. Inspired by neuroscientific findings on the association of voice and face, we propose an attention-based end-to-end neural network that learns multi-sensory associations for the task of person verification. The attention mechanism in our proposed network learns to conditionally select a salient modality between speech and facial representations that provides a balance between complementary inputs. By virtue of this capability, the network is robust to missing or corrupted data from either modality. In the VoxCeleb2 dataset, we show that our method performs favorably against competing multi-modal methods. Even for extreme cases of large corruption or an entirely missing modality, our method demonstrates robustness over other unimodal methods.
\end{abstract}
\section{Introduction}
\label{sec:intro}

From cognitive and neuroscience studies on the integration of face and voice signals in humans, it has been observed that the face-voice association is treated differently in the brain compared to other paired stimuli~\cite{von2006implicit}, and that this perceptual integration plays an important role and is actually leveraged for person recognition processing \cite{hasan2016hearing}.    
Inspired by these findings, computational models have been recently introduced to understand whether, and to what extent, such models can leverage associations between different modalities.
To investigate this multi-modal association, Nagrani et al.~\cite{nagrani2018seeing}, Horiguchi et al.~\cite{horiguchi2018face} and Kim et al.~\cite{kim2018learning} presented a face-voice cross-modal matching task by learning a shared representation for both modalities.
Neural network-based cross-modal learning is explored to distill common or complementary information from large-scale paired data.
In particular, Kim et al. showed that their computational model has similar behaviors to humans.

Based on these explorations of multi-modal computational learnability, 
we propose to investigate the use of multi-modal neural networks for the more specific and challenging task of person verification. 
There has been some work that investigates person verification using multi-modal biometric data~\cite{Choudhury1999,Luque2006,Thomas2006,hazen07,Sargin2009,Sell2018multimodal}.
These methods typically consist of independent face and voice unimodal recognition modules that are trained separately, with respective scores from the unimodal modules being combined with score fusion.
These methods also typically run in an off-line manner, whereby multiple frames of the face and several seconds of speech are used to maximize recognition performance, so there is an inherent latency built into the methodology.
On the other hand, feature-level fusion has been uncommon in the person verification.
The feature-level fusion has been more commonly adoped in audio-visual speech recognition~\cite{Neti2000, Kratt2004} from a simple concatenation of the feature to end-to-end system~\cite{Sanabria2016, Petridis2018} with synchronized audio-visual feature.
In this work, we shed light on the feature level fusion in the multi-modal person recognition.

In this paper, we explore an online audio-visual fusion system for person verification using face and voice.
In contrast to previous work on person verification, our proposed fusion method is conducted at the feature level.
In particular, we focus on the fusion of synchronized audio-visual data based on the argument that the system should naturally emphasize the time-varying contribution of each modality according to its instantaneous quality at any point time. 
Our method exploits a single video frame of the face and a short span of speech to facilitate online processing applications, while maintaining high performance relative to prior state-of-the-art.
Motivated by the attention~\cite{corbetta2002control} and the multi-sensory association mechanism of the human brain~\cite{von2006implicit}, our fusion method is implemented by an attention mechanism, such that it can learn to evaluate the salient modality of input data.
Due to the inherent robustness of this architecture, we expect stable performance even when there is corrupted information from either face or voice, due to noise masking, or missing information from basic pre-processing failures of either modality e.g., face detection, voice activity detection (VAD), etc.
We experimentally verify that this audio-visual fusion network is robust to corrupted and missing information from one modality.
We also analyze the attention layer output to see how it behaves under certain characteristics of the input.



\begin{figure*}[ht]
     \centering
     \begin{subfigure}[b]{0.3\textwidth}
         \centering
         \includegraphics[width=\textwidth]{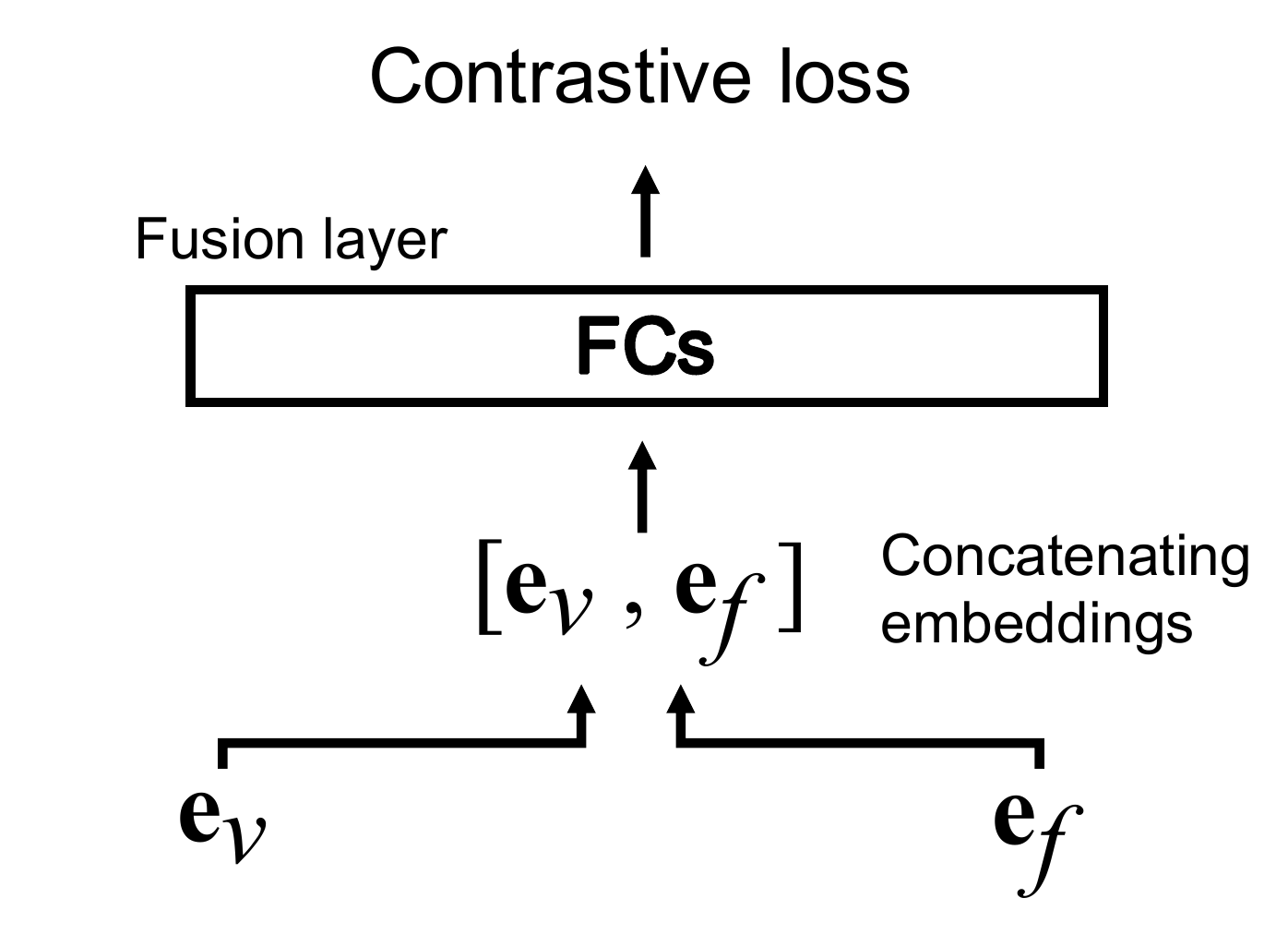}\vspace{-3mm}
         \caption{System $\mathcal{A}$}
     \end{subfigure}
     \hfill
     \begin{subfigure}[b]{0.3\textwidth}
         \centering
         \includegraphics[width=\textwidth]{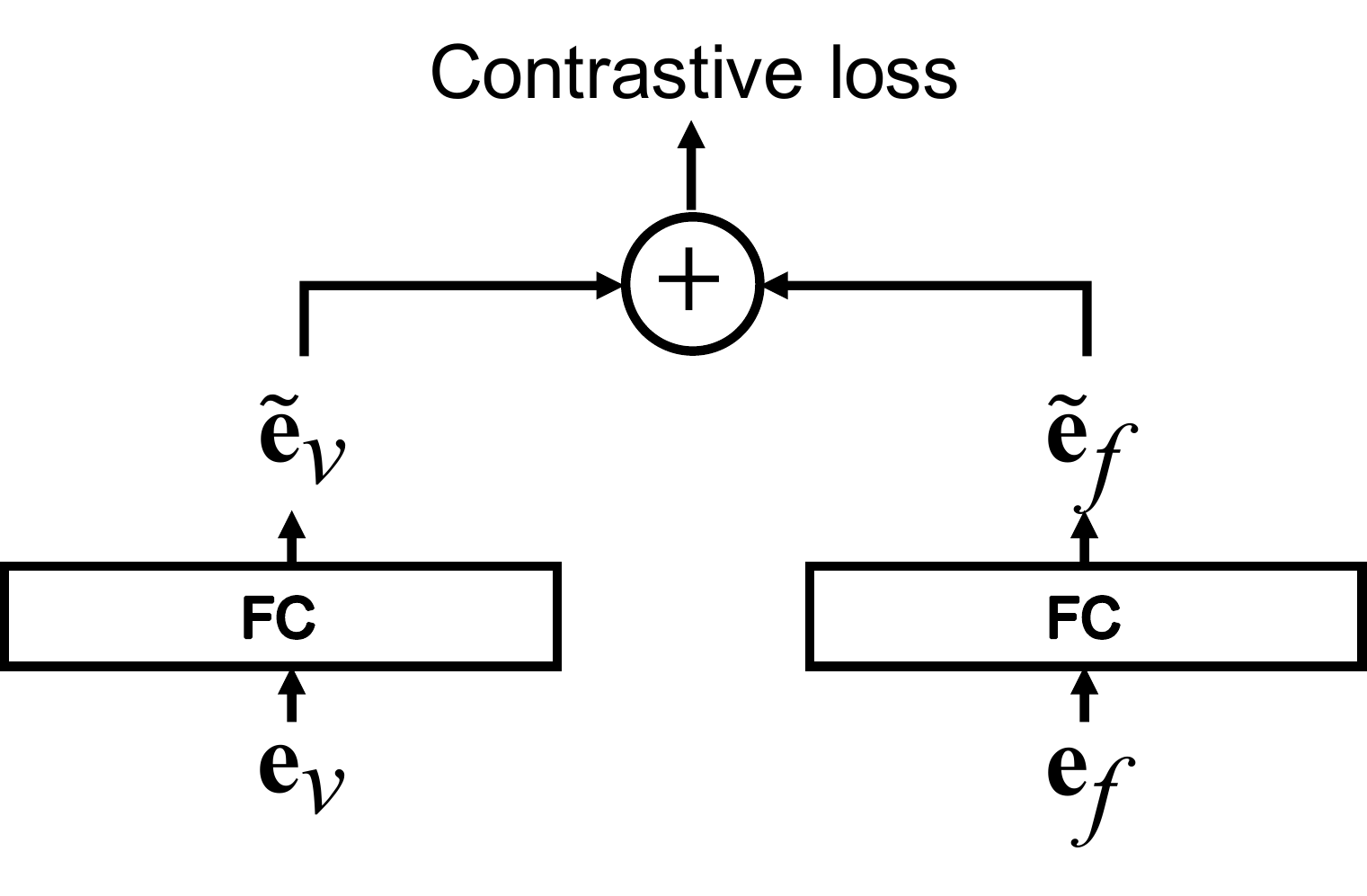}\vspace{-3mm}
         \caption{System $\mathcal{B}$}
     \end{subfigure}
     \hfill
     \begin{subfigure}[b]{0.3\textwidth}
         \centering
         \includegraphics[width=\textwidth]{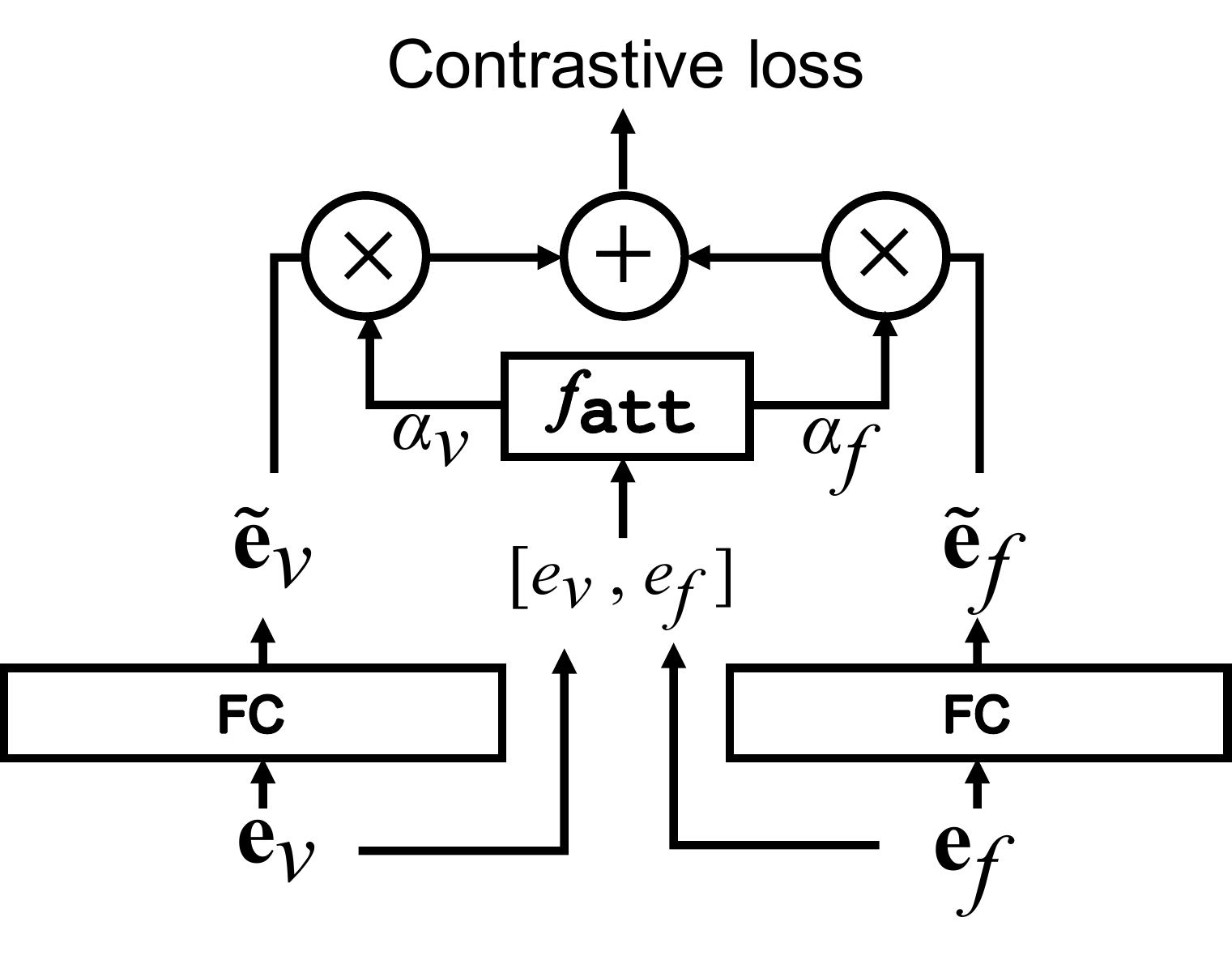}\vspace{-3mm}
         \caption{System $\mathcal{C}$}
     \end{subfigure}
        \caption{Neural network based fusion approaches. $\mathbf{e}_v$: speaker embedding, $\mathbf{e}_f$: face embedding}
        \label{fig:attention}
\end{figure*}

\section{Online Person verification from Video}
The verification of a person's identity is often achieved by using information from a single modality that contains the biometric signal, such as images for face identification and audio for speaker verification.  When multiple modalities are available, such as in video recordings of someone speaking, then opportunities exist to explore fusion of information from both modalities.  
Both vision and hearing must address challenges due to variation in a persons appearance or voice, or occlusion due to environmental conditions.  In the case of vision, the image of a person's face will appear differently due to physical changes in a person's appearance, emotional state, occlusion due to other objects, and will depend on position and orientation relative to the camera etc.  Likewise a person's voice can change due to health, or emotional state, and will be affected by environmental noise, reverberation and channel conditions.  

One interesting difference between face and speaker ID technologies is that high quality face ID can be obtained from a single image of a person's face.  In video data, this corresponds to a single instance in time, and can be sampled many times a second.  In contrast, to achieve the same level of performance for speaker verification tasks typically requires a much longer sample of speech from the talker (e.g., 10-30sec of speech are typical conditions, with a few seconds of speech being a much more challenging task).  This discrepancy is because, unlike images of faces, the speech signal is highly time-varying due to the nature of speech production.  A random snippet of speech can be dramatically different from another, even when spoken by the same talker, due to differences in the acoustic-phonetic sequences present in the samples.  The characteristics of a talkers voice are more reliably extracted when the duration of a speech recording contains more examples of the different sounds produced by the talker.  For person verification, there is some truth to the mantra that a picture is worth a thousand words!

When processing video data there will be situations where one modality or the other may be corrupted or altogether missing.  A corrupted modality can be caused by a false alarm of a pre-processing step such as face detection or voice activity detection (VAD).  For example, a face detector may incorrectly identify a face, or detect the wrong face or region in the video, or the VAD might be activated by background noise that does not contain a human voice. These corrupted inputs could easily confound a multi-modal network to the point where its performance could be worse than fusing separate unimodal systems.  
When one modality is completely missing, 
one easiest solution in practice would be to switch to apply an alternative backup unimodal system to the uncontaminated modal data.
We will demonstrate that our multi-modal system performs favorably against this systematic approach even in the complete missing case.

\section{Audio-Visual Multi-modal Fusion}
In this section, we describe the proposed multi-modal fusion approach and its voice and face representation subsystems.
Our method is distinguished from previous studies by its use of a feature-level fusion approach based on neural network models. 
Given discriminative face and speaker representations extracted from each subsystem, our attention layer evaluates each contributions of the  representations.
Then, we combine them according to the estimated contributions, so that a joint representation is obtained.
We learn this whole fusion network for the person verification task without additional supervision for the attention.
In the test phase, we compute the similarity of joint representations between the query (enrollment) and test samples to verify identities.


In the following sections we elaborate the proposed fusion approach and the speech and face sub-systems used in our experiment.

\subsection{Proposed Fusion Approach}
We develop a multi-modal attention model that can pay attention to the salient modality of inputs while producing a powerful fusion representation appropriate for the person verification task.
This is inspired by the humans' multi-sensory capability. Among diverse facets of the human multi-sensory system, the presence of the selective attention \cite{corbetta2002control} allows humans to first pick salient information even from crowded sensory inputs.
The human attention mechanism dynamically brings salient features to the forefront as needed without collapsing holistic information into blurry abstraction.

The realization of this attention mechanism in deep neural networks has achieved successes in various machine learning applications.
Our attention network is similar to the soft attention~\cite{bahdanau2014neural} which is differentiable.
While most previous work applies spatial or temporal attention, our attention is extended to be attentive across the modality axis.
Given face and speaker embeddings $\mathbf{e}_f$ and $\mathbf{e}_v$, we define the attention score ${\hat a}_{\{f,v\}}$ through attention layer $f_\texttt{att}(\cdot)$ as 
\begin{equation}
{\hat a}_{\{f,v\}} = f_\texttt{att}(\left[ \mathbf{e}_f, \mathbf{e}_v \right]) = \textbf{W}^\top\left[ \mathbf{e}_f, \mathbf{e}_v \right] + \textbf{b},
\end{equation}
where $\textbf{W} \in \mathbb{R}^{m\times d}$  and $\textbf{b} \in \mathbb{R}^{m}$ are the learnable parameters of the attention layer, $m$ and $d$ denote the number of modality to fuse and the input dimension of the attention layer respectively, and $\mathbf{e}_f$ and $\mathbf{e}_v$ will be discussed in the next subsection.
Then, the fused embedding $\mathbf{z}$ is constructed by the weighted sum as
\begin{equation}
\mathbf{z} = \sum_{i \in \{f,v\}} \alpha_i \tilde{\mathbf{e}}_i,\quad\textrm{where}\quad 
\alpha_i = \frac{\exp({\hat a}_i)}{\sum_{k \in \{f,v\}} \exp({\hat a}_k)}
, i\in\{f,v\},
\end{equation}
where $\tilde{\mathbf{e}}$ denotes the projected embeddings to a co-embedding space compatible with the linear combination.
To map $\tilde{\mathbf{e}}_{\{f,v\}}$ from ${\mathbf{e}}_{\{f,v\}}$, we used a Fully Connected (FC) layer with 600 hidden nodes, i.e. $\tilde{\mathbf{e}} \in \mathbb{R}^{600}$. We do not used non-linearity in the FC layer. 
We train the attention networks by the contrastive loss on the joint embedding $\mathbf{z}\in \mathbb{R}^{600}$.
For each training step, we used 60 positive and negative pairs, a total of 120 pairs for each mini-batch, and all pairs were sampled from the VoxCeleb2 development set.

The proposed attention networks allow us to naturally deal with corruption or missing data from either modality.
In our framework, the attention networks spontaneously learn to assess the quality of given multi-modal data implicitly.
For example, if the audio signal is largely corrupted by surrounding noise, the attention network would switch off the voice representation path and only rely on the face representation, and vice versa.
In this way, as long as at least one modality provides appropriate information for the task, this model will be able to perform person verification. 

\vspace{1mm}\noindent\textbf{Relationship with Other Fusion Methods}\quad
In the context of the multi-modal person verification, the traditional score-level fusion with logistic regression has been investigated up to these days~\cite{Choudhury1999,Luque2006,Thomas2006,hazen07,Sargin2009,Sell2018multimodal}.
These score fusion methods do not leverage any large capacity deep neural networks which are capable of dealing with non-trivial fusion strategy.
One can come up with an extension based on the above approaches, where FC layers are stacked on top of the concatenated speaker and face embeddings, $\mathbf{e}_v$ and $\mathbf{e}_f$, as shown in Figure~\ref{fig:attention}-(a), i.e. System $\mathcal{A}$. We used 2 FC layers with 1,200 and 600 hidden nodes and ReLUs for non-linearities in the first FC layer. 
This can be regarded as a feature level fusion similar to Nagrani et al.~\cite{nagrani2018seeing}.
A downside of this would be the fact that the performance of the system is degraded by corrupted modal data.

Another neural network based fusion can be accomplished as shown in Figure~\ref{fig:attention}-(b). 
FC layers are stacked on top of respective embeddings, $\mathbf{e}_v$ and $\mathbf{e}_f$, without a nonlinear activation function. 
This layer simply projects each modality embeddings into a joint audio-visual subspace.  
Then, the projected embeddings, $\widetilde{\mathbf{e}}_v$ and $\widetilde{\mathbf{e}}_f$, are combined by the summation operation, and used for the contrastive loss function as we did.
The summation based ensemble considers both modalities contribute equally, typically yielding a mean representation which can be easily biased with a large contamination~\cite{candes2011robust}.

Our method adaptively estimates the weights of each embedding to construct a joint representation. Either of weight can be turned off if the embedding would degrade the end performance.
This feature is not only robust but also able to deal with missing or a large corruption of the data.

\subsection{Voice and Face Representations}
To obtain discriminative embeddings for face and voice, $\mathbf{e}_f$ and $\mathbf{e}_v$, 
we exploit the existing deep neural network based representations.

\vspace{1mm}\noindent\textbf{Voice embedding}\quad
Voice embeddings generally exploit a large dataset including augmented data with added background noise. 
A voice embedding can be extracted from one of the hidden layers from a neural network trained to classify $N$ speakers in the training dataset. In a previous study, we proposed a frame-level voice embedding to extract robust speaker information by modifying the DNN structure after training is complete~\cite{Shon2018frame}. For training, the VoxCeleb1 development dataset was used. 
Details can be found in~\cite{Shon2018frame} since we used the same system.
Frame-level voice embeddings are extracted every 10ms using a 25ms frame window. 
Before fusion, a total of 10 and 100 successive voice embeddings are averaged to create a voice embedding which spans 115ms and 1015ms, respectively since a single frame-level voice embedding spanning 25ms is too short to extract voice characteristics reliably. 

\vspace{1mm}\noindent\textbf{Face embedding}\quad
Our face embeddings are extracted by using  FaceNet~\cite{Schroff2015} pre-trained on CASIA-WebFace.\footnote{\url{https://github.com/davidsandberg/facenet}\\ We used this reproduced open model, which has been improved by the maintainers with several modifications.
The modifications include the dimension change of the last layer from 128-D to 512-D. We use the last 512-D \texttt{FC7} layer activation of this FaceNet version as the face embedding.}
Since the provided face region annotations in the VoxCeleb datasets are coarse, 
we re-align and crop faces by the face and landmark detectors in Dlib.\footnote{\url{http://dlib.net}}

\section{Experiments}
In this section, we evaluate the proposed method with various baselines. 
In Sec.~\ref{sec:exp_fusion}, we compare the person verification performance with several multimodal fusion approaches as well as unimodal methods in the ordinary scenario that both modal data is given.
Then, we demonstrate the robustness of the proposed method against corrupted data in Sec.~\ref{sec:exp_corrupt}.
Moreover, we analyze the behavior of the attention layer according to interpretable attributes, including head pose and facial appearance traits, in Sec.~\ref{sec:exp_analysis}.

\subsection{Experimental Environment}
\label{sec:exp_env}
For our experiments, 
we used the VoxCeleb1 \& 2 datasets~\cite{Nagraniy2017,Chung2018}, which include multimedia data with a reliable pre-processing step to obtain face regions and voice segments.
VoxCeleb1 \& 2 have more than 1,281,352 utterances from 7,365 speakers and both datasets have development and test set splits.
For verification performance measurement, we made a test trial set using the VoxCeleb2 Test set which contains 36,693 video clips from 120 speakers. 
We made 300 positive trials (i.e., the same speaker from different clips) and 300 negative trials (i.e., different speaker) trials per speaker, for a total of 71,790 trials.\footnote{The number is slightly less than 72,000 because there are a few individuals who have less than five video clips.}. We used cosine similarity to measure the distance of two embeddings.

Voice and face embeddings were extracted in 600 and 512 dimension respectively. For training the fusion network ($\mathcal{A},\mathcal{B},\mathcal{C}$), we extracted 1 frame per second and its relevant audio segment with 0.115 sec and 1.015 sec. Both embeddings were L2-normalized to have unit length before feeding into the fusion network. To test, we extract a single frame and its relevant audio segment randomly in each video clip. Thus, a total of 36,693 still images and 0.115 sec (or 1.015 sec) audio segments are used for the test trials. The performance was measured in terms of Equal Error Rate (EER) and minimum Detection Cost Function (mDCF) ($P_\textit{target}=0.01$)~\cite{sre16}. 

\subsection{Fusion Performance}
\label{sec:exp_fusion}
As shown in Table~\ref{tab:fusion}, the voice embedding shows significantly worse performance than the face embedding. This is natural because we only use 0.115 sec, 1.015 sec which is a very short segment to extract reliable representations from text-independent speech.  Score-level fusion was done using logistic regression by calibrating on the VoxCeleb2 development set~\cite{Brummer2006}. The system $\mathcal{A, B}$ and $\mathcal{C}$ show neural network-based fusion approaches. While system $\mathcal{A}$ and $\mathcal{B}$ shows slightly better performance than the score-level fusion on EER, System $\mathcal{C}$ show a notable gain in both EER and mDCF.


\begin{table}[t]
\centering
\resizebox{0.4\textwidth}{!}{%
\begin{tabular}{l|cc|cc}
\hlineB{2}
 & \multicolumn{2}{c|}{$l$=0.115 sec} & \multicolumn{2}{c}{$l$=1.015 sec} \\ \hline
Systems & EER & mDCF & EER & mDCF\\ \hlineB{2}
Voice embedding ($\mathbf{e}_v$) & 41.27 & 0.999 & 14.50 & 0.863 \\ 
Face embedding ($\mathbf{e}_f$)& 8.03 & 0.631 & 8.03 & 0.631\\ \hline
Score-level fusion & 7.83 & 0.623 & 5.78 & 0.491 \\ 
System $\mathcal{A}$& 7.74 & 0.634 & 5.52 & 0.478 \\ 
System $\mathcal{B}$& 7.81 & 0.625 & 5.56 & 0.472\\ 
System $\mathcal{C}$ (Proposed) & \textbf{7.46} &\textbf{ 0.611} & \textbf{5.29} &\textbf{ 0.456} \\ \hlineB{2}
\end{tabular}%
}
\caption{Person verification performance on VoxCeleb2 test set. $l$ is a length of audio segment to extract voice embedding.}
\label{tab:fusion}
\end{table}

\subsection{Effect on Corrupted and Missing Modality}
\label{sec:exp_corrupt}
To see the performance under a corrupted or missing modality of either voice and face, we generated random noise drawn from a standard normal distribution and zero vector. Random noise mimics embeddings from an corrupted modality from an image without a face or audio without a voice due to an error in the pre-processing step. The zero vector is for the case of a  missing modality and this can be easily handled by switching the multi-modal system to unimodal system. However, we were interested in the scenario where we only used a single universal system and measured the performance when either modality did not exist. In table~\ref{tab:missing}, the proposed system $\mathcal{C}$ shows better performance for both the corrupted and missing modality condition by assessing the quality of the embedding in the attention layer.

\begin{table}[t]
\centering
\begin{subtable}[t]{0.45\textwidth}
\centering
\resizebox{1\linewidth}{!}{%
\setlength{\tabcolsep}{2.5pt}
\begin{tabular}{c|cc|cc|cc|cc}
\hlineB{2}
 & \multicolumn{4}{c|}{Voice null embeddings} & \multicolumn{4}{c}{Face null embeddings} \\ \cline{2-9} 
 & \multicolumn{2}{c|}{Random} & \multicolumn{2}{c|}{Zeros} & \multicolumn{2}{c|}{Random} & \multicolumn{2}{c}{Zeros} \\ \hline
Systems & EER & mDCF & EER & mDCF & EER & mDCF & EER & mDCF \\ \hlineB{2}
Score fusion & 8.05 & 0.633 & 8.03 & \textbf{0.631} & 49.99 & 0.999 & 41.27 & 0.999 \\ 
System $\mathcal{A}$ & 8.51 & 0.712 & 7.59 & 0.648 & 38.81 & 0.999 & 35.51 & 0.999 \\ 
System $\mathcal{B}$ & 8.76 & 0.748 & 7.51 & 0.637 & 37.74 & 0.999 & \textbf{34.12} & 0.999 \\ \hline
System $\mathcal{C}$
 & \multirow{ 2}{*}{\textbf{7.77}} & \multirow{ 2}{*}{\textbf{0.626}} & \multirow{ 2}{*}{\textbf{7.50}} & \multirow{ 2}{*}{0.633} & \multirow{ 2}{*}{\textbf{37.23}} & \multirow{ 2}{*}{0.999} & \multirow{ 2}{*}{34.22} & \multirow{ 2}{*}{0.999} \\ 
 (Proposed) & & & & & & & & \\
 \hlineB{2}
\end{tabular}%
}
\caption{$l= 0.115$ sec}
\end{subtable}
\vspace{5pt}

\begin{subtable}[t]{0.45\textwidth}
\centering
\resizebox{1\linewidth}{!}{%
\setlength{\tabcolsep}{2.5pt}
\begin{tabular}{c|cc|cc|cc|cc}
\hlineB{2}
 & \multicolumn{4}{c|}{Voice null embeddings} & \multicolumn{4}{c}{Face null embeddings} \\ \cline{2-9} 
 & \multicolumn{2}{c|}{Random} & \multicolumn{2}{c|}{Zeros} & \multicolumn{2}{c|}{Random} & \multicolumn{2}{c}{Zeros} \\ \hline
Systems & EER & mDCF & EER & mDCF & EER & mDCF & EER & mDCF \\ \hlineB{2}
Score fusion & 8.19 & 0.634 & 8.03 & \textbf{0.631} & 28.18 & 0.995 & 14.5 & \textbf{0.863} \\ 
System $\mathcal{A}$ & 8.64 &	0.732&	7.64&	0.649&	15.42&	0.960&	13.27&	0.897 \\ 
System $\mathcal{B}$ & 8.69&	0.724&	\textbf{7.61}&	0.647&	16.52&	0.970&	14.55&	0.901 \\ \hline
System $\mathcal{C}$
 & \multirow{ 2}{*}{\textbf{7.89}} & \multirow{ 2}{*}{\textbf{0.623}} & \multirow{ 2}{*}{7.65} & \multirow{ 2}{*}{0.636} & \multirow{ 2}{*}{\textbf{12.64}} & \multirow{ 2}{*}{\textbf{0.905}} & \multirow{ 2}{*}{\textbf{12.23}} & \multirow{ 2}{*}{0.871} \\ 
 (Proposed) & & & & & & & & \\
 \hlineB{2}
\end{tabular}%
}
\caption{$l= 1.015$ sec}
\end{subtable}

\caption{Performance under corrupted and missing modality on either voice and face. $l$ is a length of audio segment to extract voice embedding.}
\label{tab:missing}
\end{table}

Interestingly, the neural network based fusion systems, particularly the proposed fusion approach, obtain better performance than using a unimodal embedding, even for the case that information is only partially available.
In the neuroscience study, it has been observed that unimodal perception gets a benefit from the multisensory association of ecologically valid and sensory redundant stimulus pair ~\cite{Kreiegstein2006multisensory}.
As an extension of this observation, we can interpret as the fusion network learns the association of the multisensory data, and it becomes available to extract more robust feature even without multisensory data.



\subsection{Analysis of the Attention Layer}
\label{sec:exp_analysis}
We analyze the behavior of the attention layer in our networks.
In order to parse what information it has learned and its behavior according to interpretable attributes, we conduct control experiments with facial appearance attributes.

By measuring probabilities of face/voice attention weights conditioned by an attribute in the test set, we investigate the existence of the statistical correlation between the attribute and the attention, and its tendency.
We obtain the attributes of the VoxCeleb2 test set by using the state-of-the-art, Rude et al.~\cite{rudd2016moon} and Feng et al.~\cite{feng2018prn} for 40 facial appearance attributes~(defined in the CelebA dataset~\cite{Liu2015}) and 3D head orientation, respectively.
We focus on the relationship between \emph{the behavior of attention weights} and attributes, considering the fact that Kim et al. \cite{kim2018learning} already showed the connections of face/voice representations with certain demographic attributes. 

As a statistical measure, given an attribute $A$, we measure the expectation of the probability $\mathbb{E} P(\alpha_f {>} {\bar \alpha}_f|A{=}\textrm{true})$, ${\bar \alpha}_f$ denotes the global mean of the face attention over all the test data, and likewise for the voice.
Since the probability estimate follows the expectation of the Bernoulli trial, we measure the statistical significance by 95$\%$ binomial proportion (Wald) confidence interval.
While the attribute estimation methods have extremely low-failure rate profile, taking into account subtle outlier effects, we conservatively regard the 95$\%$-confidence lower bound estimates as being a significant signal if greater than 60$\%$ (greater than the random chance).

From Table~\ref{tab:headorientation}, we could not find any correlation between head orientation and attention weights. We postulate that the FaceNet embedding is learned to be sufficiently head orientation invariant, so the attention layer turns out to be insensitive to the quality of the embedding according to the orientation. Table~\ref{tab:attribute} shows the 7 attributes that their lower bound is above the 60\%. It is interesting that, in the case that a person is with the temporary attributes, such as ``Wearing Hat,'' ``Sideburns,'' ``Goatee'' and ``Mustache'', the fusion system is likely to concentrate on the voice with a much higher chance than random. 
Also, the very strong attribute like ``Bald,'' ``Blond hair'' and ``Straight Hair'' shows correlation with attention weights.

\begin{table}[t]
\centering
\begin{subtable}[ht]{1\linewidth}
\centering
\resizebox{1\textwidth}{!}{%
\begin{tabular}{ c|cc|cc|cc}
\hlineB{2}
Head orientation & \multicolumn{2}{c|}{ $|\theta|$\textless30\textdegree} & \multicolumn{2}{c|}{30\textdegree\textless$|\theta|$\textless60\textdegree} & \multicolumn{2}{c}{60\textdegree\textless$|\theta|$} \\ \hline
 & V (\%) & F (\%) & V (\%) & F (\%) & V (\%) & F (\%) \\ \hline
Yaw & 43 & 57 & 46& 54 & 44 & 56 \\ 
Pitch & 44 & 56 & 41 & 59 & 42 & 58 \\ 
Roll & 44 & 56 & 43 & 57 & 47 &53 \\ \hlineB{2}
\end{tabular}%
}
\caption{Head orientation attributes. (V: voice, F: face)}
\label{tab:headorientation}
\end{subtable}
\vspace{5pt}

\begin{subtable}[ht]{1\linewidth}
\centering
\resizebox{0.7\textwidth}{!}{%
\begin{tabular}{c|c|c|c}
\hlineB{2}
Facial Attibutes & Voice (\%) & Face (\%) & 95\% C.I. \\ \hline
Bald & \textbf{74.89} & 25.11 & $\pm$ 4.02 \\ 
Blond Hair & 32.17 & \textbf{67.83} & $\pm$ 1.51 \\
Goatee & \textbf{70.06} & 29.94 & $\pm$ 1.38 \\ 
Mustache & \textbf{72.96} & 27.04 & $\pm$ 1.73 \\ 
Sideburns & \textbf{65.60} & 34.40 & $\pm$ 1.81 \\ 
Straight Hair & 29.65 & \textbf{70.35} & $\pm$ 1.09 \\ 
Wearing Hat & \textbf{72.62} & 27.38 & $\pm$ 2.14 \\ \hlineB{2}
\end{tabular}%
}
\caption{Facial appearance attributes}
\label{tab:attribute}
\end{subtable}
\caption{The expectation of $ P(\alpha_v {>} {\bar \alpha}_v|A{=}\textrm{true})$ and $ P(\alpha_f {>} {\bar \alpha}_f|A{=}\textrm{true})$, where $A$ denotes attributes.
C.I. stands for the (Wald) confidence interval.
For head orientation, the front face is represented by all the angle of yaw, pitch and roll equal to 0\textdegree.
}
\end{table}



\section{Conclusion}

Motivated from the recent studies about the multi-modal association, we proposed a feature-level attentive fusion network for audio-visual online person verification task. 
The temporally synced face image and voice segment assumption encourages the network to learn about the quality of the embedding to verify a person's identity. 
The learned embeddings of both modalities share a compatible space (co-embedding space) by virtue of the simple linear combination rule to obtain the fused representation. 
Besides the better performance than the traditional score-level fusion, it has a large advantage to handle the severe condition such as the presence of the corrupted and missing modality. 
The attention mechanism is also analyzed to understand the correspondence between attention weights and interpretable attributes of visual perception.
In addition to visual appearance traits, it would be interesting to further investigate the attention behavior in terms of speech characteristics, such as pitch, language, dialect, etc, as a future direction. 

\clearpage
\newpage

\bibliographystyle{IEEEbib_r1}
\small
\bibliography{refs}

\end{document}